# Evaluating prose style transfer with the Bible

Keith Carlson[1], Allen Riddell[3] and Daniel Rockmore[1,2,4]

[1]Department of Computer Science, and [2]Department of Mathematics, Dartmouth College Hanover, NH 03755, USA
[3]School of Informatics and Computing, Indiana University Bloomington, Bloomington, IN 47405, USA
[4]The Santa Fe Institute, 1399 Hyde Park Road, Santa Fe, NM 87501, USA

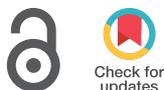 KC, 0000-0002-3273-5293; AR, 0000-0002-4967-0879

In the prose style transfer task a system, provided with text input and a target prose style, produces output which preserves the meaning of the input text but alters the style. These systems require parallel data for evaluation of results and usually make use of parallel data for training. Currently, there are few publicly available corpora for this task. In this work, we identify a high-quality source of aligned, stylistically distinct text in different versions of the Bible. We provide a standardized split, into training, development and testing data, of the public domain versions in our corpus. This corpus is highly parallel since many Bible versions are included. Sentences are aligned due to the presence of chapter and verse numbers within all versions of the text. In addition to the corpus, we present the results, as measured by the BLEU and PINC metrics, of several models trained on our data which can serve as baselines for future research. While we present these data as a style transfer corpus, we believe that it is of unmatched quality and may be useful for other natural language tasks as well.

## 1. Introduction

Written prose is one way in which we communicate our thoughts to each other. Given a 'message', there are many ways to write a sentence capable of conveying the embedded information, even when they are all written in the same language. Sentences can communicate essentially the same information but do so using different 'styles'. That is, the various versions may have essentially the same meaning or semantic content, and insofar as they use different words are each 'paraphrases' of each other. These paraphrases, while sharing the same semantic content, are not necessarily interchangeable. When writing a sentence we frequently consider not only the semantic content we wish to communicate, but also the manner, or style, in which we express it. Different wording may convey different levels of politeness or familiarity with the reader, display different cultural information about the writer, be easier to understand for certain populations, etc., *Style transfer*, or stylistic paraphrasing, is the task of rewriting a sentence such that we preserve the meaning but alter the style.

The problem of style transfer is clearly relevant for the creation of natural language generation systems. The translations, paraphrases, summarizations, and other language generated by a natural language system are only useful if the outputs are understood and accepted by the intended audience. This may require us to target certain levels of simplicity, formality, or other characteristics of style in the language produced.

There are many features of the prose which contribute to the perceived style of a text including sentence length, use of passive or active voice, vocabulary level, tone and level of formality. Analysis and classification of style focusing on sentiment, usage of stop-words, formality, etc., have all been the subject of study [1–4]. Similarly, generation of language targeting one aspect of style such as simplicity, formality, length and use of active voice have received attention [5–8], and some work has even been done to try to control several of these properties simultaneously [9]. Fewer results exist which do not consider any of these aspects explicitly, but instead use a more general view of style. These systems generally require parallel data for training and testing their results and parallel style transfer corpora are in short supply. A few recent exceptions use a corpus of Shakespearean plays and their modernizations for the task [10,11]. Even more recently, researchers have published results of targeting a style using only unlabelled training data. Some of these systems generate text not conditioned on an input and cannot be applied to this problem [12], and others only show results on a sentiment transfer task [13,14]. Generally, unsupervised systems still need parallel data for evaluation and may benefit from some amount of parallel data during training. We believe that one of the major barriers to automatic style transfer research is the relatively low amount of high-quality parallel data.

The main contributions of our work are as follows.

## 1.1. Identification of a highly parallel corpus

Style transfer can naturally be viewed as a machine translation problem where the source language and target language are simply different textual styles. The style transfer task is treated this way in much of the existing work on style transfer and related problems [10,11,15,16]. Despite this obvious connection, many breakthroughs in machine translation systems have not been directly applied to style transfer. As many previous authors have noted, the major difficulty seems to lie in finding a suitably large parallel corpus of different styles [17–20].

Bibles, with their well-demarcated sentence and verse structure provide such a corpus. Herein we identify a novel and highly parallel corpus useful for the style paraphrasing task: 34 stylistically distinct versions of the Bible (Old and New Testaments). Each version is understood as embodying a unique writing style. The versions in this corpus were created with a wide range of intentions. Versions such as the Bible in Basic English were written to be simple enough to be understood by people with a very limited vocabulary. Other versions, like the King James Version, were written centuries ago and use very distinctive archaic language. In addition to being viewed individually, the versions can also be partitioned according to different stylistic criteria, any one of which could be a goal of a paraphrasing. For example, metrics that enable the identification of versions deemed 'simple' could identify a subcorpus that would allow training towards the task of text simplification. Versions identified as using 'old' language could be used to train towards the task of 'text archaification'. Such richly parallel datasets are difficult to find, but this corpus provides such a wide range of text that it could be used to focus on a variety of stylistic features already present within the data. While many parallel corpora require alignment before they can be used, here verse numbers immediately identify equivalent pieces of text. Thus, in these data the text has all been aligned by humans already. This eliminates the need to use text alignment algorithms which may not produce alignments that match human judgement. Our work splits books of the Bible into training, testing and development sets. We then publish these sets using all eight of the publicly available Bible versions in our more complete corpus and list the versions we use which are not public. This easy to access and free to use, standardized, parallel corpus (and the accompanying benchmarks we produce) is the main contribution of our work.

We hope that the publication of such a corpus will lead to immediate application of some machine translation techniques that were previously not applicable to style transfer, and that over time these techniques can be fine-tuned to better handle the nuanced differences between machine translation and style transfer.

There are some recent results employing systems which do not require a parallel corpus for training at all, both in machine translation [21,22] and stylistic paraphrasing [13,20]. While these results are encouraging, the methods are intended to be used in situations where a parallel corpus is not available. The unsupervised techniques are not, at this time, meant to outperform supervised methods in cases when suitable paired training data are available. Furthermore, the results of these models are still evaluated using parallel data. Even if textual style transfer research becomes more focused on unsupervised learning methods, there will be a need for parallel text representing different styles for testing purposes.

## 1.2. Publication of baseline results for our corpus

To showcase the usefulness of this corpus, and to facilitate future comparison, we train several baseline models and publish the results. We train and evaluate encoder–decoder recurrent neural networks (Seq2Seq) and Moses [23], a statistical machine translation system. We report the BLEU [24] and PINC [15] scores of the outputs of these systems and provide the textual outputs themselves to allow research using other metrics to be easily compared with these baselines.

# 2. Related work

## 2.1. Style transfer datasets

Ours is clearly not the first parallel dataset created for style transfer, and the existing datasets have their own strengths and weaknesses.

One of the most used style transfer corpora was built using articles from Wikipedia and Simple Wikipedia to collect examples of sentences and their simplified versions [25]. These sources further were used with improved sentence alignment techniques to produce another dataset which included classification of each parallel sentence pair's quality [26]. More recently, word embeddings were used to inform alignment and yet another Wikipedia simplification dataset was released [17].

The use of Wikipedia for text simplification has been criticized generally, and some of the released corpora denounced for more specific and severe issues with their sentence alignments [27]. The same paper also proposed the use of the Newsela corpus for text simplification. These data consist of 1130 news articles, each professionally rewritten four times to target different reading levels.

A new dataset targeting another aspect of style, namely formality, should soon be made publicly available [18]. The Grammarly's Yahoo Answers Formality Corpus (GYAFC) was constructed by identifying 110 000 informal responses containing between 5 and 25 words on Yahoo Answers. Each of these was then rewritten to use more formal language by Amazon Mechanical Turk workers.

While these datasets can all be viewed as representing different styles, simplicity and formality are only two aspects of a broader definition of style. The first work to attempt this more general problem introduced a corpus of Shakespeare plays and their modern translations for the task [10]. This corpus contains 17 plays and their modernizations from http://nfs.sparknotes.com and versions of eight of these plays from http://enotes.com. While the alignments appear to mostly be of high quality, they were still produced using automatic sentence alignment which may not perform the task as proficiently as a human. The larger sparknotes dataset contains about 21 000 aligned sentences. This magnitude is sufficient for the statistical machine translation methods used in their paper, but is not comparable to the corpora usually employed by neural machine translation systems.

Most of these existing parallel corpora were not created for the general task of style transfer [17,18,25–27]. A system targeting only one aspect of style may use techniques specific to that task, such as the use of simplification-specific objective functions [19]. So while we can view simplification and formalization as types of style transfer, we cannot always directly apply the same methods to the more general problem.

The Shakespeare dataset [10], which does not focus on only simplicity or formality, still contains only two (or three if each modern source is considered individually) distinct styles. Standard machine translation corpora, such as WMT-14 (http://www.statmt.org/wmt14/translation-task.html), have parallel data across many languages. A multilingual corpus not only provides the ability to test how generalizable a system is, but can also be leveraged to improve results even when considering a single source and target language [28].

Some of these existing corpora require researchers to request access to the data [18,27]. Access to high-quality data are certainly worth this extra step, but sometimes response times to these requests can be slow. We experienced a delay of several months between requesting some of these data and receiving it. With the current speed of innovation in machine translation, such delays in access to data may make these corpora less practical than those with free immediate access.

## 2.2. Machine translation and style transfer models

As mentioned, style transfer has obvious connections to work in traditional language-to-language translation. The Seq2Seq model was first created and used in conjunction with statistical methods to perform machine translation [29]. The model consists of a recurrent neural network acting as an encoder, which produces an embedding of the full sequence of inputs. This sentence embedding is then used by another recurrent neural network which acts as a decoder and produces a sequence corresponding to the original input sequence.

Long short-term memory (LSTM) [30] was introduced to allow a recurrent neural network to store information for an extended period of time. Using a formulation of LSTM which differs slightly from the original [31], the Seq2Seq model was adapted to use multiple LSTM layers on both the encoding and decoding sides [32]. This model demonstrated near state-of-the-art results on the WMT-14 English-to-French translation task. In another modification, an attention mechanism was introduced [33] which again achieved near state-of-the-art results on English-to-French translation.

Other papers proposed versions of the model which could translate into many languages [33,34], including one which could translate from many source languages to many target languages, even if the source–target pair was never seen during training [28]. The authors of this work make no major changes to the Seq2Seq architecture, but introduce special tokens at the start of each input sentence indicating the target language. The model can learn to translate between two languages which never appeared as a pair in the training data, provided it has seen each of the languages paired with others. The idea of using these artificially added tags was applied to related tasks such as targeting level of formality or use of active or passive voice in produced translations [6,7].

This work on machine translation is relevant for paraphrase generation framed as a form of monolingual translation. In this context, statistical machine translation techniques were used to generate novel paraphrases [35]. More recently, phrase-based statistical machine translation software was used to create paraphrases [36].

Tasks such as text simplification [5,16] can be viewed as a form of style transfer, but generating paraphrases targeting a more general interpretation of style was first attempted in 2012 [10]. All of these results employed statistical machine translation methods.

The advances mentioned previously in neural machine translation have only started to be applied to general stylistic paraphrasing. One approach proposed the training of a neural model which would 'disentangle' stylistic and semantic features, but did not publish any results [37]. Another attempt at text simplification as stylistic paraphrasing is [38]. They generate artificial data and show that the model performs well, but do no experiments with human-produced corpora. The Shakespeare dataset [10] recently was used with a Seq2Seq model [11]. Their results are impressive, showing improvement over statistical machine translation methods as measured by automatic metrics. They experiment with many settings, but in order to overcome the small amount of training data, their best models all require the integration of a human-produced dictionary which translates approximately 1500 Shakespearean words to their modern equivalent.

# 3. Data

## 3.1. Data collection

As stated above, a significant contribution of this paper is the identification and publication of Bible versions as a stylistic paraphrasing dataset. For our work, we collected 33 English translations of the Bible from www.biblegateway.com, and also the Bible in Basic English from www.o-bible.com. We found that seven of these collected versions are in the public domain and thus can be freely distributed. Additionally, the Lexham English Bible (http://www.lexhamenglishbible.com/) has a permissive licence which allows it to be distributed for free. These eight public versions are used to create the corpus that we release. Other versions can be acquired relatively easily and inexpensively, but may not be distributed due to prevailing copyright law. Table 1 displays the complete list of versions.

These Bible versions are highly parallel and high-quality, having been produced by human translators. Sentence-level alignment of parallel text is needed for many NLP tasks. Work exists on methods to automatically align texts [25,26,39], but the alignments produced are imperfect and some have been criticized for issues which decrease their usefulness [27]. The Bible corpus is human-aligned by virtue of the consistent use of books, chapters and verses across translations. While many verses are single sentences some are sentence fragments or several sentences. This is not problematic as we only require the parallel text to be aligned in small parts which have the same meaning, but there is no obvious reason that this must be at a strict sentence level.

**Table 1.** Names of publicly available Bible versions and other versions we used followed by their standard abbreviations in parenthesis. Text was collected from www.biblegateway.com (and BBE from www.o-bible.com).

| public domain Bible versions | other versions used |
| --- | --- |
| Bible in Basic English (BBE) | New Life Version (NLV) |
| World English Bible (WEB) | New International Reader's Version (NIRV) |
| Young's Literal Translation (YLT) | International Children's Bible (ICB) |
| Lexham English Bible (LEB) | Easy-To-Read Version (ERV) |
| Douay-Rheims 1899 American Edition (DRA) | New Century Version (NCV) |
| American Standard Version (ASV) | Contemporary English Version (CEV) |
| Darby Translation (DARBY) | Good News Translation (GNT) |
| King James Version (KJV) | God's Word Translation (GWT) |
| | Names of God Bible (NOG) |
| | Jubilee Bible 2000 (JUB) |
| | New King James Version (NKJV) |
| | Modern English Version (MEV) |
| | English Standard Version (ESV) |
| | 1599 Geneva Bible (GNV) |
| | New International Version (NIV) |
| | Holman Christian Standard Bible (HCSB) |
| | 21st Century King James Version (KJ21) |
| | New Living Translation (NLT) |
| | New Revised Standard Version (NRSV) |
| | Common English Bible (CEB) |
| | New English Translation (NET) |
| | International Standard Version (ISV) |
| | Revised Standard Version (RSV) |
| | New American Bible Revised Edition (NABRE) |
| | The Living Bible (TLB) |
| | The Message (MSG) |

Some Bible versions contain instances of several verses combined to one. For example, we may find a Bible version with 'Genesis 1:1-4' instead of singular instances of each of the four verses. We remove these aggregated verses from our data to keep the alignment more fine-grained and consistent. There are over 31 000 verses in the Bible, so even with this regularization we still have over 1.7 million potential source and target verse pairings in the publicly available data and over 33 million pairs in the full dataset.

## 3.2. Data splitting

We need to split our data into training, testing and development sets. We do so by selecting entire Bible books to be included in each set to ensure that the text in the training data are not too similar to anything that appears in testing or development. Additionally, we expect that the language used in the New Testament may differ from that used in the Old Testament. We therefore want to ensure that each of our splits contains books from each of them. The test data are constructed by selecting two random Old Testament books and two random New Testament books, and the process is repeated for the development data. In the published data, the testing set books are Judges, 1 Samuel, Philippians and Hebrews. The development set books are 1 Kings, Zephaniah, Mark and Colossians. The remaining

**Table 2.** BLEU scores between full text of Bible versions. The verses of the version of each row are treated as the candidates and the column version's verses are treated as the reference.

|       | YLT   | DARBY | KJV   | WEB   | DRA   | LEB   | BBE   | ASV   |
|-------|-------|-------|-------|-------|-------|-------|-------|-------|
| YLT   | 100   | 26.43 | 23.61 | 19.33 | 13.57 | 15.15 | 9.42  | 25.87 |
| Darby | 26.46 | 100   | 52.79 | 37.86 | 23.94 | 22.44 | 16.27 | 55.49 |
| KJV   | 23.89 | 53.38 | 100   | 41.04 | 30.18 | 19.6  | 17.76 | 68.72 |
| WEB   | 19.78 | 39.49 | 41.24 | 100   | 20.37 | 30.07 | 19.15 | 53.11 |
| DRA   | 16.3  | 29.39 | 35.56 | 23.69 | 100   | 17.76 | 15.29 | 31.64 |
| LEB   | 17.89 | 26.71 | 22.72 | 33.67 | 17.46 | 100   | 18.49 | 25.98 |
| BBE   | 11.72 | 20.35 | 21.8  | 22.59 | 15.49 | 19.4  | 100   | 22.75 |
| ASV   | 26.48 | 56.84 | 69.09 | 53.01 | 26.95 | 22.51 | 18.72 | 100   |

**Table 3.** Pairings of Bible versions created. For each pairing, parallel training, test and development files are created and number of lines in each are reported.

| source versions | target versions | # train lines | # dev lines | # test lines |
|-----------------|-----------------|---------------|-------------|--------------|
| all             | all             | 28 693 558    | 1 707 252   | 1 920 108    |
| public          | public          | 1 534 582     | 91 780      | 102 732      |
| public          | ASV             | 192 414       | 11 456      | 12 843       |
| public          | BBE             | 192 324       | 11 481      | 12 843       |
| BBE             | ASV             | 27 584        | 1637        | 1835         |
| KJV             | ASV             | 27 608        | 1637        | 1835         |
| YLT             | BBE             | 27 595        | 1642        | 1835         |

35 Old Testament books and 23 New Testament books are used as the training split. The verse numbers which are at the beginning of each line are removed as they are always identical for each pair of verses and so make no interesting contribution.

In addition to releasing the data with all public versions, we release files with the training, testing and development text targeting only a single version. These are useful for training when the system being trained is not capable of multi-style paraphrasing (such as Moses) or when making comparisons to such a system. To decide on good target versions, we looked at the BLEU scores between the full text of every pair of versions. The results of this analysis on the public Bibles can be seen in table 2.

Some versions are highly similar according to the BLEU metric but some are quite different. For example, treating ASV as a candidate and KJV as the reference has a score of 69.09 but comparing BBE to YLT only gives a score of 11.72. Because of this we would expect a system trained to transfer ASV text into the KJV style to outscore one trained for the BBE to YLT task. We want to consider situations where there is relatively little modification required to the input as well as those which will need drastic stylistic revision. We pick ASV as a single-version target because, when treating all other public versions as a candidate, it has the highest average BLEU score when treated as a reference. Similarly, we will use BBE as a target since it has the lowest average BLEU score. In addition to creating splits using all public versions as sources, and BBE and ASV as targets, we want to investigate models' performance-specific version→version pairs of varying similarities. To this end, we also create parallel files of only KJV to ASV (easy), BBE to ASV (hard) and YLT to BBE (very hard). These files can be used to train models on their own, or used to test those which were trained using all public versions for the source and the correct single version as a target. The version pairings for which we create and publish (excluding all→all) parallel training, testing and development files can be seen in table 3.

The full verse-aligned texts of all public Bible versions are available on github (https://github.com/keithecarlson/StyleTransferBibleData). Additionally, we publish the public version parallel testing, training and development files discussed so that future work with this corpus can make use of standardized data. We include code and a walkthrough of our process for all of the baseline results which we report.

# 4. Baseline models

## 4.1. Moses

The statistical machine translation system Moses [23] is an established baseline for testing new paraphrasing corpora and models. To use it this way, one recasts paraphrasing as a monolingual translation task on the paired data as mentioned above. Previous such uses include the work of Chen & Dolan [15], Xu *et al.* [10] and Wubben *et al.* [36], who found that it outperformed paraphrasing based on word substitution. It was also used as a baseline in [11] for stylistic paraphrasing of Shakespeare into present-day English.

Moses is designed to translate from a single source language to a single target language and in the previous uses of Moses for style transfer only two distinct styles were used. Since our corpus contains examples of many styles, the proper choice of training data for Moses is not as obvious. We could give Moses training examples using all versions as sources and all versions as targets and it should learn to produce good paraphrases, but we want the paraphrases produced by Moses to be in the style of a specific version. Even when targeting only a single style we have the option to use many versions as source sentences during training or only a single source version. To explore both of these options fully we use each of the single-target parallel files discussed above, those with single source versions and those with multiple, to train Moses models. We train Moses five times in total, once each using the (public versions → ASV), (public versions → BBE), (KJV → ASV), (BBE → ASV) and (YLT → BBE) parallel files and then use these models to decode all test sets with the appropriate target version. For example, the Moses model trained on (public versions → ASV) is evaluated on the (public versions → ASV), (KJV → ASV) and (BBE → ASV) test sets.

For all runs of Moses, we use mgiza for word alignment [40], kenlm for the language model [41] and mert [40] to fine-tune the model parameters to the development dataset. All of these tools are provided with Moses. The language model built is of order 5.

## 4.2. Seq2Seq

Encoder–decoder recurrent neural networks (Seq2Seq) have been widely used and adapted for machine translation in recent years [29,32,33,42]. One such paper introduced artificial tags at the beginning of each source example to indicate the language to target during decoding [28]. This minor change allowed the model to perform multilingual and zero-shot translation. These models generally require a large number of training examples to produce high-quality results.

Application of Seq2Seq models to stylistic paraphrasing has only just started to be explored. We are aware of only one existing work [11] which uses such a network in this context on real data. To overcome the relatively small corpus, the authors [11] use a human-expert produced dictionary giving the translations of Shakespearean words into modern equivalents.

We use our new, and relatively large Bible corpus to train Seq2Seq models. Our corpus contains many versions, and the number of training examples when using a single version as a source and a single version as a target is small. To fully take advantage of how richly parallel our data are, we use the tagging trick of [28]. In each source verse, we prepend a tag indicating the version to be targeted. For example, if the target style for a verse pair is that of the American Standard Version, we start off the source sentence with an '<ASV>' token. Using this method we are able to train a separate Seq2Seq model using each of the following parallel version pairs (all versions → all versions), (public versions → public versions), (public versions → ASV) and (public versions → BBE). We experimented with running this model on the single source and single target files, such as (YLT→BBE), but the results were poor because the amount of training data was too small for this model.

The Seq2Seq model requires a fixed vocabulary which contains the tokens which will be encountered by the model. Names and rare words are often difficult to handle in NLG tasks. Sometimes they are replaced with a generic 'Unknown' token [33,43,44]. Recently, byte pair encoding was used to create a vocabulary of subword units which removes the need for such a token [45]. In the resulting vocabulary, rare words are not present but the smaller units which make up the word are. For example, in our data the rarely seen word 'ascribed' is replaced by the more frequent subwords 'as' and 'scribed'. We generated a vocabulary of the 30 000 most frequent subword units from the training portion of all of our Bible versions. This vocabulary was then applied to each of the samples by replacing any word which was not in the vocabulary with its constituent subword units.

We use a multi-layer recurrent neural network encoder and multi-layer recurrent network with attention for decoding. This set-up is similar to those described by Sutskever *et al.* [32] and Bahdanau *et al.* [33].

The encoder and decoder each have three layers of size 512 and use LSTM [31] and dropout between layers [46], with probability of dropping set to 0.3. Each uses a trainable embedding layer to project each token into a 512-dimensional vector before it is passed to the LSTM layers. The encoder is bi-directional [33,47] and has residual connections between the layers [48]. The decoder uses an attention mechanism as described by Bahdanau *et al.* [33] to focus on specific parts of the output from each step of the encoder. The exact software and configuration of our model can be found on github (https://github.com/keithecarlson/StyleTransferBibleData).

During training, mini-batches of 64 verse-pairs are randomly selected from the training corpus. Each of the target and source sentences are truncated to 100 tokens if necessary. The model's parameters are adjusted using the 'Adam optimiser' [49]. A checkpoint of the model is saved periodically[1] during training. The checkpoints are all evaluated on the development data and the one with the lowest loss is selected.

During inference a single source sentence is fed into the model but the target sentence is not provided. Unlike during training, the decoder is fed its own prediction as input for the next timestep. The decoder performs a beam search [32] with a width of 10 to produce the most likely paraphrase.

# 5. Experiments

As indicated above, we train Moses and Seq2Seq models on a variety of source-target pairings of our Bible Corpus. Our Seq2Seq model is implemented using a publicly available library [50] which itself makes use of the API provided by Tensorflow [51]. See figure 1 for an overview of our entire work process.

The code and data to run the experiments which use only the publicly available portion of our data are available (https://github.com/keithecarlson/StyleTransferBibleData).

## 5.1. Metrics

For evaluation of the models, we use several established measures. We first calculate BLEU [24] scores for all results. BLEU is a metric for comparing parallel corpora which rewards a candidate sentence for having n-grams which also appear in the target. Although it was created for evaluation of machine translation, it has been found that the scores correlated with human judgement when used to evaluate paraphrase quality [36]. The correlation was especially strong when the source sentence and candidate sentence differed by larger amounts as measured by Levenshtein distance over words.

BLEU gets at some of what a good paraphrase should accomplish (similarity), but a good (i.e. interesting) paraphrase should use different words than the source sentence, as noted by Chen & Dolan [15]. They introduce the PINC score which 'computes the percentage of n-grams that appear in the candidate sentence but not in the source sentence'. The PINC score makes no use of target sentence, but rewards a candidate for being dissimilar from the source. To capture a candidate's similarity to the target and dissimilarity from the source, they use both the BLEU and PINC scores together. They find that BLEU scores correlate with human judgement of semantic equivalence and that PINC scores correlated highly with human ratings of lexical dissimilarity. Lexical dissimilarity on its own is important for paraphrasing, but as previously mentioned, a high lexical dissimilarity may also strengthen the correlation of BLEU scores and human judgement of paraphrase quality [36]. We will use and report both PINC and BLEU for evaluation as can be found in previous work on stylistic paraphrasing [10,11].

While these metrics have been widely and previously used for paraphrase evaluation, many other measures have been proposed for style transfer and related problems [10,19,20]. By publishing not only PINC and BLEU scores, but the full texts produced by our systems themselves, we allow future researchers to compare to our baselines using any of these existing or newly proposed metrics.

## 5.2. Results

For each of the single target test sets, we identify the Moses model and the Seq2Seq model which achieves the highest BLEU score. The results of the evaluation metrics for these models' outputs can be seen in table 4. The results of all models and test sets can be found in our github.

---

[1]We found that models trained on smaller amounts of data tend to overfit faster. To ensure we have a high-quality checkpoint we need to save them more frequently when training on the smaller datasets. We saved a checkpoint every 5000 steps on public→public and all→all training and every 1000 steps when using only a single version as a target.

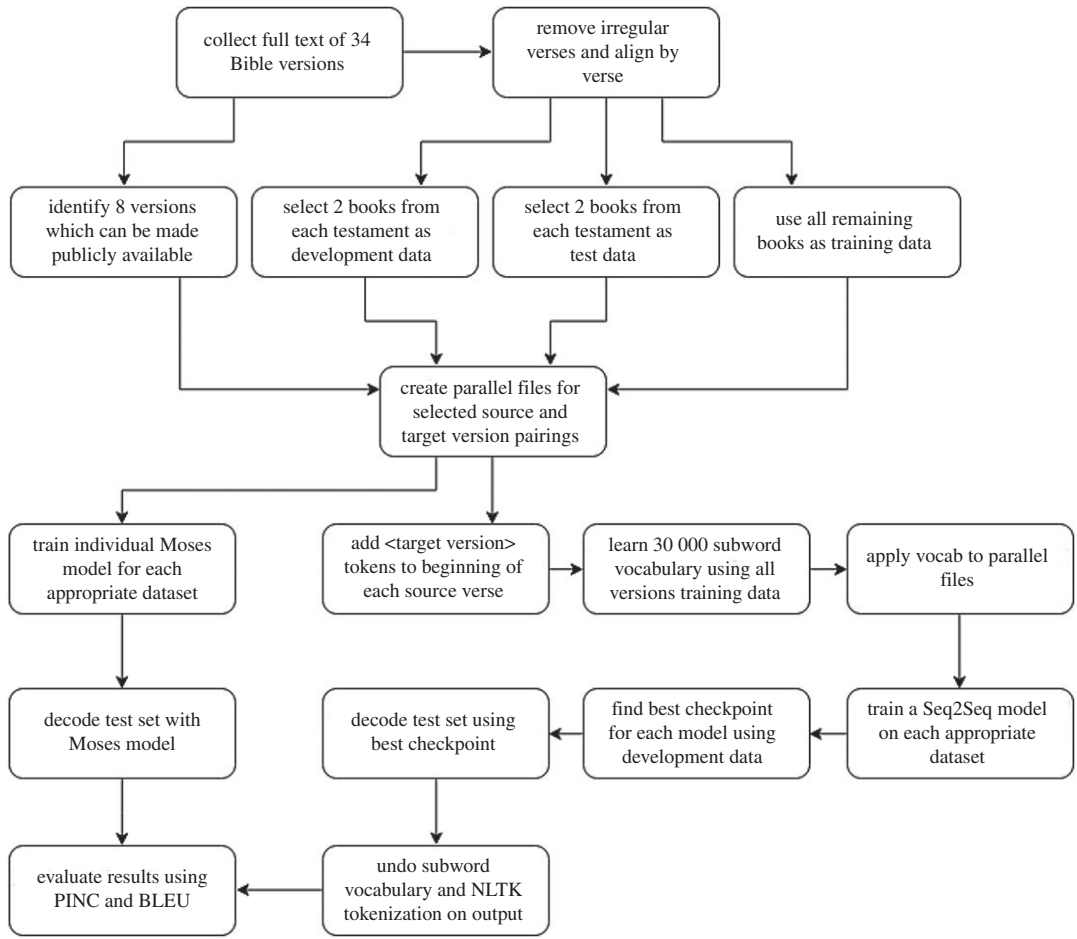

**Figure 1.** A diagram of the experimental workflow.

**Table 4.** The BLEU and PINC scores of the best Moses and best Seq2Seq models for each test set. The best model is defined here as the one which achieves the highest BLEU score on the test set. One BLEU score and One PINC score in each row are in italics. These represent the best score on the test set by each of these metrics.

| test set | best moses training | best Seq2Seq training | Moses BLEU | Seq2Seq BLEU | Moses Pinc | Seq2Seq PINC |
|---|---|---|---|---|---|---|
| KJV→ASV | KJV→ASV | public→ASV | *71.16* | 65.61 | 16.84 | *24.68* |
| BBE→ASV | BBE→ASV | public→public | *31.28* | 30.17 | 47.03 | *56.8* |
| YLT→BBE | YLT→BBE | public→public | 24.01 | *24.94* | 66.47 | *72.51* |
| public→ASV | public→ASV | public→ASV | *46.87* | 46.00 | 28.67 | *42.22* |
| public→BBE | public→BBE | public→public | *30.03* | 29.71 | 56.49 | *64.54* |

We note a few trends across all of these results. First, we find that the Seq2Seq model always has a higher PINC score than the Moses model. This matches our qualitative observation that Moses makes changes to the source sentence less aggressively than Seq2Seq and often leaves large portions of the input unchanged. A few examples showing this can be seen in table 5.

Moses always performed best when it was trained on the exact source and target versions that it would eventually be tested on. Seq2Seq, however, seems to often benefit from additional versions being included. For example, when decoding the YLT→BBE test set the best Moses was trained using the YLT→BBE training set. The best Seq2Seq model, however, was trained using the training set where all public versions were used as both the source and target. As mentioned above, we experimented with training Seq2Seq on single source version and single target version data, such as YLT→BBE, but the results were poor. In this case though the Seq2Seq model trained on public→public also outperformed the one trained on public→BBE and on all→all. It seems the increase in number of training examples going from

**Table 5.** Examples of Moses and Seq2Seq output which showcase Moses making fewer modifications to the source sentence.

| source | Moses output | Seq2Seq output |
| --- | --- | --- |
| And that there may not be any evil liver, or any man without respect for God, like Esau, who let his birthright go for a plate of food. | And that there shall not be any evil liver, or any man without respect unto God, as Esau that let his birthright go for a plate of food. | And the liver shall not sin, nor the transgression of God, like Esau; who let his birthright go for the dish. |
| His legs were covered with plates of brass and hanging on his back was a javelin of brass. | His legs were covered with flakes of brass and hanged on his shoulder was a javelin of brass. | And he overlaid his legs with plates of brass, and put on his back a spear of brass. |

**Table 6.** Examples which show Moses and Seq2Seq outputs which have correctly made changes to better imitate the target style or which have missed some obvious part of the transfer.

| source | target | Moses output | Seq2Seq output |
| --- | --- | --- | --- |
| (BBE) Then Samuel gave him an account of everything, keeping nothing back. And he said, It is the Lord; let him do what seems good to him. | (ASV) And Samuel told him every whit, and hid nothing from him. And he said, It is Jehovah: let him do what seemeth him good. | Then Samuel told him of all things not. And he said, It is Jehovah; let him do that which seemeth him good. | And Samuel told all things, and did not hold back. And he said, It is Jehovah; let him do what seemeth good to him. |
| (YLT) And Jehovah saith, 'Judah doth go up; lo, I have given the land into his hand'. | (BBE) And the Lord said, Judah is to go up: see, I have given the land into his hands. | And the Lord said, 'Judah will go up, see, I have given the land into his hand.' | And the Lord said, Judah will go up; see, I will give the land into his hand. |

public→BBE to public→public was helpful for decoding into BBE even when none of the extra examples were targeting BBE. This is similar to results which found that adding additional languages to a multilingual Seq2Seq model could improve performance [28]. This improvement seems to be bounded in our case, however, as the public→public model also outperforms the all→all model when decoding YLT→BBE. It is possible that when using all 34 Bible versions, there is simply too much for our formulation of the model to learn. Since it has to be able to produce 34 different styles it may not be able to focus on perfecting each of them. It could also be that the non-public Bible versions have some characteristics which make them ultimately detrimental to use as training for a model which will be evaluated on the YLT→BBE task.

In four out of the five pairings evaluated Moses outperforms Seq2Seq as measured by BLEU. It is only on the YLT→BBE test set that the best Seq2Seq model achieves a higher BLEU score than the best Moses model. This test set is the most demanding of the model, as the source and target sentences are the least similar, as can be seen in table 2. This performance seems related to our earlier observation that Moses is more conservative in making changes to the source. In situations where relatively little modification to the source is required, such as in the KJV→ASV task, the auto-encoding tendencies of Moses can be quite helpful. When more drastic revision is required, however, the more aggressive tendencies of Seq2Seq begin to become more effective.

Qualitatively, both models seem to be changing text to better imitate the targeted style. In the first example in table 6, Moses and Seq2Seq correctly use the archaic verb 'seemeth' and replace 'Lord' with 'Jehovah'. Both modifications are stylistically correct when targeting ASV. We notice some cases where Moses seems to be unable to pick up on relatively simple stylistic markers. In the second example of table 6, we see a translation from YLT to BBE. YLT uses quotation marks when someone is speaking, but BBE does not. While the text produced by both Moses and Seq2Seq resembles BBE in style, Moses has not removed the quotation marks which were present in the source. This is despite the fact that the Moses model was only trained with BBE targets and so has never actually seen a training example where it should have produced quotation marks.

# 6. Conclusion and future work

In this paper, we collected a large, previously untapped dataset of aligned parallel text, Bible translations. We present this corpus for the task of prose style transfer.

Existing corpora used for style transfer target only a single aspect of style (such as simplicity), are much smaller, have been criticized for problems in their alignment which reduce usefulness, require potentially lengthy waiting periods for access or contain only two styles.

By contrast, our data use a broad understanding of style, with each Bible version embodying a unique style. Since each version contains over 31 000 verses, we are able to produce a training set of over 1.5 million unique pairings of source and target verses from the eight publicly available versions. The data are human aligned by virtue of the consistent use of book, chapter and verse number across all versions, eliminating the risk of alignment errors by automatic methods. Finally, the public Bible versions, and our splits of them into training, testing and development data are freely and immediately available through our public repository (https://github.com/keithecarlson/StyleTransferBibleData).

While we present the corpus for style transfer, it is rare to find data that is human aligned and so richly parallel. These qualities may make our corpus useful for a variety of other natural language tasks as well. For example, the large number of aligned translations in these data could prove useful for training towards the traditional paraphrasing task in which a specific style is not targeted. Alternatively, researchers could choose some aspect of style, such as simplicity or formality, and partition the corpus based on that criteria. The partitioned corpus could then be used to train models which produce text with the desired characteristic.

Style transfer can naturally be viewed as monolingual machine translation, but lack of an appropriate training corpus has made the direct application of many machine translation models difficult. To showcase the usefulness of our data and establish baseline results for future research, we train statistical machine translation models and encoder–decoder recurrent neural networks on our corpus. We find that these two systems perform similarly on most of the decoding datasets. In general, Moses performs slightly better, achieving a higher BLEU score on four of our five evaluations. This superiority is increased when the task requires less modification of the source sentence to match the target. Seq2Seq makes gains on Moses when the task requires more aggressive editing of the source, and is able to outperform Moses on the most demanding of our five tests.

It is likely that some previously published modifications to Seq2Seq would result in immediate performance improvements. Candidates from the machine translation literature include: coverage modelling [52] to help track which parts of the source sentence have already been paraphrased and the use of a pointer network [53] to allow copying of words directly from the source sentence. Pointer networks have already been used for style transfer [11], and seem likely to be useful for our multi-style corpus as well.

While application of state-of-the-art neural machine translation methods may yield near-term improvements in style transfer, we expect that at some point the most successful models will need to treat this problem as a separate task. A large parallel corpus will allow researchers to explore both similarities as well as the nuanced differences between the tasks of style transfer and machine translation. We hope that these data inspire the creation of style-transfer-specific architectures.

Data accessibility. Data, code and a walk-through to run our experiments on the public domain portion of our data are available at https://github.com/keithecarlson/StyleTransferBibleData. Copyright law prevents distribution of many of the versions of the Bible we used. The repository contains only the eight Bible versions which may be freely distributed.

Authors' contributions. K.C., A.R. and D.R. designed the experiment, interpreted results and wrote, reviewed, edited and approved the final paper. K.C. collected the data, wrote the code and ran the experiments.

Competing interests. We declare we have no competing interests.

Funding. No funding has been received for this article.